\documentclass[preprint,authoryear,12pt]{elsarticle}
\usepackage{mathrsfs}
\usepackage{graphics,graphicx,epsfig}
\usepackage{amsmath,mathrsfs,amssymb,amsfonts}
\usepackage{algorithm,algorithmic}
\usepackage{multirow,bm}

\newtheorem{theorem}{Theorem}
\newtheorem{lemma}{Lemma}
\newtheorem{definition}{Definition}

\newtheorem{corollary}{Corollary}

\DeclareMathOperator{\sgn}{sgn}
\newenvironment{proof}{\par\noindent{\bf Proof\ }}{\hfill\BlackBox\\[2mm]}

\def\x{{\bm x}}
\def\X{{\mathcal{X}}}
\def\Y{{\mathcal{Y}}}
\def\D{{\mathcal{D}}}
\def\B{{\mathcal{B}}}

\journal{Artificial Intelligence Journal}

\begin{document}

\begin{frontmatter}

\title{On the Consistency of AUC Pairwise Optimization}

\author{Wei Gao and Zhi-Hua Zhou\corref{cor1}}

\address{National Key Laboratory for Novel Software Technology\\
             Nanjing University, Nanjing 210023, China}
\cortext[cor1]{Email: zhouzh@lamda.nju.edu.cn}
\begin{abstract}
AUC (area under ROC curve) has always been an important evaluation criterion popularly used in diverse learning tasks such as class-imbalance learning, cost-sensitive learning, learning to rank and information retrieval. Many learning approaches are developed to optimize AUC, whereas owing to its non-convexity and discontinuousness, most approaches work with pairwise surrogate losses such as exponential loss, hinge loss, etc; therefore, an important theoretic problem is to study on the AUC consistency based on minimizing pairwise surrogate losses.

In this paper, we introduce the generalized calibration for AUC optimization, and prove that the generalized calibration is necessary yet insufficient for AUC consistency. We then provide a new sufficient condition for the AUC consistency of learning approaches based on minimizing pairwise surrogate losses, and from this finding, we prove that exponential loss, logistic loss and distance-weighted loss are consistent with AUC. In addition, we derive the \textit{$q$-norm hinge loss} and \textit{general hinge loss} that are consistent with AUC. We also derive the regret bounds for exponential loss and logistic loss, and present the regret bounds for more general surrogate losses in the realizable setting. Finally, we prove regret bounds that disclose the equivalence between the pairwise exponential surrogate loss of AUC and the exponential surrogate loss of accuracy, and one direct consequence of such finding is the equivalence between AdaBoost and RankBoost in the limit of infinite sample.
\end{abstract}


\end{frontmatter}

\section{Introduction}\label{sec:intro}
AUC (Area Under ROC Curve) is an important evaluation criterion, which has been adopted in diverse learning tasks such as cost-sensitive learning, class-imbalance learning, learning to rank, information retrieval, etc. \citep{Elkan2001,Freund:Iyer:Schapire:Singer2003,Cortes:Mohri2004,Balcan2007,Ailon:Mohri2008,Clemencon:Vayatis2009,Clemencon:Vayatis:Depecker2009,Kotlowski:Dembczynski:Hullermeier2011,Flach:Orallo:Ramirez2011}, where traditional criteria such as \textit{accuracy}, \textit{precision}, \textit{recall}, etc. are inadequate \citep{Provost:Fawcett:Kohavi1998,Provost:Fawcett2001} since AUC is irrelevant to class distribution.

Owing to the non-convexity and discontinuousness, it is not easy, or even infeasible, to optimize AUC directly since such optimization often yields NP-hard problems. To make a compromise for avoiding computational difficulties, pairwise surrogate losses that can be optimized more efficiently are usually adopted in practical algorithms, e.g., exponential loss \citep{Freund:Iyer:Schapire:Singer2003,Rudin:Schapire2009}, hinge loss \citep{Brefeld:Scheffer2005,Joachims2005,Zhao:Hoi:Jin:Yang2011}, least square loss \citep{Gao:Jin:Zhu:Zhou2013}, etc.

An important theoretic problem is how well does minimizing such convex surrogate losses lead to improving the actually AUC; in other words, does the expected risk of learning with surrogate losses converge to the Bayes risk of AUC? Consistency (also called Bayes consistency) guarantees that optimizing a surrogate loss will yield an optimal function with Bayes risk in the limit of infinite sample. Thus, the above problem, in a formal expression, is whether the optimization of surrogate losses is consistent with AUC.

\subsection{Our Contribution}
We first introduce the generalized calibration for AUC optimization based on minimizing the pairwise surrogate losses, and find that the generalized calibration is necessary yet insufficient for AUC consistency. For example, hinge loss and absolute loss are calibrated but inconsistent with AUC. The deep reason is that, for pairwise surrogate losses, minimizing the expected risk over the whole distribution is not equivalent to minimizing the conditional risk on each pair of instances.

We then provide a new sufficient condition for the AUC consistency of learning approaches based on minimizing pairwise surrogate losses. From this finding, we prove that exponential loss, logistic loss and distance-weighted loss are consistent with AUC. In addition, we derive the \textit{$q$-norm hinge loss} and \textit{general hinge loss} that are consistent with AUC. We also derive the regret bounds for exponential loss and logistic loss, and present the regret bounds for more general surrogate losses in the realizable setting.

Finally, we provide regret bounds that disclose the equivalence between the pairwise exponential surrogate loss of AUC and the exponential surrogate loss of accuracy; in other words, the exponential surrogate loss of accuracy is consistent AUC, while the pairwise surrogate loss of AUC is consistent with accuracy by selecting a proper threshold. One direct consequence of such finding is the equivalence between AdaBoost and RankBoost in the limit of infinite sample.

\subsection{Related Work}
The studies on AUC can be traced back to 1970's in signal detection theory \citep{Egan1975}, and it has been widely used as a criterion in medical area and machine learning \citep{Provost:Fawcett:Kohavi1998,Provost:Fawcett2001,Elkan2001}. In model selection, AUC also exhibits better measure than accuracy theoretically and empirically \citep{Huang:Ling2005}. AUC can be estimated under parametric \citep{Zhou:Obuchowski:Mcclish2002}, semi-parametric \citep{Hsieh:Turnbull1996} and non-parametric \citep{Hanley:McNeil1982} assumptions, and the non-parameteric estimation of AUC is popularly applied in machine learning and data mining, equivalent to the Wilcoxon-Mann-Whitney (WMW) statistic test of ranks \citep{Hanley:McNeil1982}. In addition, \citet{Hand2009} and \citet{Flach:Orallo:Ramirez2011} present the incoherent and coherent explanations of AUC as a measure of aggregated classifier performance, respectively.

AUC has always been regarded as an performance measure for information retrieval and learning to rank, especially for bipartite ranking \citep{Cohen:Schapire:Singer1999,Freund:Iyer:Schapire:Singer2003,Cortes:Mohri2004,Rudin:Schapire2009,Rudin2009}. Various Generalization bounds are presented to understand the prediction beyond the training sample \citep{Agarwal:Graepel:Herbrich:Har-Peled:Roth2005,Usunier:Amini:Gallinari2005,Cortes:Mohri:Rastogi2007,Clemencon:Lugosi:Vayatis2008,Agarwal:Niyogi2009,Rudin:Schapire2009,Wang:Khardon:Pechyony:Jones2012,Kar:Sriperumbudur:Jain:Karnick2013}. In addition, the learnability of AUC has been studied in \citep{Agarwal:Roth2005,Gao:Zhou2013}.

Consistency is an important theoretic issue in machine learning. For example, \citet{Breiman2004} showed that exponential loss converges to the Bayes classifier for arcing-style greedy boosting algorithms, and \citet{Buhlmann:Yu2003} proved the consistency of boosting algorithms with respect to least square loss. \cite{Lin2002} and \cite{Steinwart2005} studied the consistency of support vector machines. For binary classification, \citet{Zhang2004ann} and \citet{Bartlett:Jordan:McAuliffe2006} provided the most fundamental and comprehensive analysis, and many famous algorithms such as boosting, logistic regression and SVMs are proven to be consistent. Further, the consistency studies on multi-class learning and multi-label learning have been addressed in \citep{Zhang2004,Tewari:Bartlett2007} and in \citep{Gao:Zhou2011,Gao:Zhou2013a}, respectively. Also, it is well-studied on the consistency of learning to rank \citep{Clemencon:Lugosi:Vayatis2008,Cossock:Zhang2008,Xia:Liu:Wang:Zhang:Li2008,Xia:Liu:Li2009,Duchi:Mackey:Jordan2010}.

In contrast to previous studies on consistency \citep{Zhang2004,Zhang2004ann,Bartlett:Jordan:McAuliffe2006,Tewari:Bartlett2007,Gao:Zhou2011,Gao:Zhou2013a} that focused on single instances, our work concerns about the pairwise surrogate losses over a pair of instances from different classes. Such difference yields that previous consistent analysis is sufficient to study on conditional risk whereas our analysis has to consider the whole distribution, because as to be shown in Lemma~\ref{lem2}, minimizing the expected risk over the whole distribution is not equivalent to minimizing the conditional risk. This is a challenge for the study on AUC consistency based on minimizing pairwise surrogate losses.

\cite{Clemencon:Lugosi:Vayatis2008} formulated the ranking problems in statistical framework and achieved faster rates of convergence under noise assumptions based on new inequalities. They also studied the consistency of ranking rules, whereas our work studies the consistency of score function based on pairwise surrogate loss. This yields the fact that calibration has been shown as a necessary and sufficient condition in \citep{Clemencon:Lugosi:Vayatis2008}, whereas we will show that calibration is necessary yet insufficient condition, e.g., hinge loss and absolute loss are calibrated but inconsistent with AUC (as to be shown in Section~\ref{sec:mainres}).

\citet{Duchi:Mackey:Jordan2010} studied the consistency of supervised ranking, but it is quite different from our work. Firstly, the problem settings are different: they considered ``instances'' consisting of a query, a set of inputs and a weighted graph, and the goal is to order the inputs according to the weighted graph; yet we consider instances with positive or negative labels, and the goal is to rank positive instances higher than negative ones. Further, they established inconsistency for the logistic loss, exponential loss and hinge loss even in low-noisy setting, yet our work shows that the logistic loss and exponential loss are consistent but hinge loss is inconsistent.

\citet{Kotlowski:Dembczynski:Hullermeier2011} studied the AUC consistency based on minimizing univariate surrogate losses (e.g., exponential loss and logistic loss), and it has been generalized to a broad class of proper (composite) losses by \citet{Agarwal2013} with simpler techniques. These two studies focused on univariate surrogate losses, whereas our work considers pairwise surrogate losses that have been popularly used in many literatures \citep{Freund:Iyer:Schapire:Singer2003,Brefeld:Scheffer2005,Joachims2005,Rudin:Schapire2009,Zhao:Hoi:Jin:Yang2011,Gao:Jin:Zhu:Zhou2013}.

\subsection{Organization}
Section~\ref{sec:Pre} makes some preliminaries. Section~\ref{sec:mainres} shows that generalized calibration is necessary yet insufficient for AUC consistency, and presents a new sufficient condition with consistent surrogate losses. Section~\ref{sec:bounds} presents regret bounds for exponential loss and logistic loss, as well as regret bounds for general surrogate losses under the realizable setting. Section~\ref{sec:rel} discloses the equivalence between the exponential surrogate losses of AUC and accuracy. Section~\ref{sec:pf} presents detailed proofs and Section~\ref{sec:com} concludes this work.

\section{Preliminaries}\label{sec:Pre}
Let $\X$ be an instance space and $\Y=\{+1,-1\}$ is the label set. We denote by $\D$ an unknown (underlying) distribution over $\X\times\Y$, and $\D_\X$ represents the instance-marginal distribution over $\X$. Further, we denote $p=\Pr[y=+1]$ and conditional probability $\eta(\x)=\Pr[y=+1|\x]$. It is trivial to study the case $p=1$ (all positive instances) and $p=0$ (all negative instances), and we assume $0<p<1$ throughout this work.

For a score function $f\colon\X\to\mathbb{R}$, the AUC w.r.t. the distribution $\D$ is given by
\[
\text{AUC}_\D(f)=E[I[(y-y')f(\x)-f(x')>0]+\tfrac{1}{2} I[f(\x)=f(\x')] | y\neq y']
\]
where $(\x,y)$ and $(\x',y')$ are drawn identically and independently according to distribution $\D$, and $I[\cdot]$ is the indicator function which returns $1$ if the argument is true and $0$ otherwise. Maximizing the AUC is equivalent to minimizing the expected risk
\begin{eqnarray}
\lefteqn{R(f)=E_{(\x,y),(\x',y')\sim\D}[\ell(f,(\x,y),(\x',y')) | y\neq y']\nonumber}\\
&=&\frac{1}{2p-2p^2}E[\eta(\x)(1-\eta(\x')) \ell(f,\x,\x')+\eta(\x')(1-\eta(\x))\ell(f,\x',\x)]\label{eq:truerisk}
\end{eqnarray}
where expectation takes on $\x$ and $\x'$ drawn i.i.d. from distribution $\D_\X$, and $\ell(f,(\x,y),(\x',y'))=I[(y-y')f(\x)-f(x')>0]+\tfrac{1}{2} I[f(\x)=f(\x')]$ is also called \textit{ranking loss}. It is easy to obtain $\text{AUC}_\D(f)+R(f)=1$. Denote by the Bayes risk $R^*=\inf_{f}[R(f)]$ where the infimum takes over all measurable functions. By simple calculation, we can get the set of optimal functions as
\begin{eqnarray}
\B&=&\{f\colon R(f)=R^*\}\nonumber\\
&=&\{f\colon(f(\x)-f(\x'))(\eta(\x)-\eta(\x'))>0 \text{ if }\eta(\x)\neq\eta(\x')\}.\label{eq:AUCBayset}
\end{eqnarray}

It is easy to find that the ranking loss $\ell$ is non-convex and discontinuous, and thus a direct optimization often leads to NP-hard problems. In practice, surrogate losses that can be optimized with efficient algorithms are usually adopted. For AUC, a commonly-used formulation is given based on pairwise surrogate losses as follows:
\[
\Psi(f,\x,\x')=\phi(f(\x)-f(\x')),
\]
where $\phi$ is a convex function, e.g., exponential loss $\phi(t)=e^{-t}$ \citep{Freund:Iyer:Schapire:Singer2003,Rudin:Schapire2009}, hinge loss $\phi(t)=\max(0,1-t)$ \citep{Brefeld:Scheffer2005,Joachims2005,Zhao:Hoi:Jin:Yang2011}, least quare loss $\phi(t)=(1-t)^2$ \citep{Gao:Jin:Zhu:Zhou2013}, etc.

For pairwise surrogate loss, we define the expected $\phi$-risk as
\begin{multline}\label{eq:phi-risk}
R_\phi(f)=\frac{1}{2p(1-p)}E_{\x,\x'\sim\mathcal{D}^2_{\mathcal{X}}}[\eta(\x)(1-\eta(\x')) \phi(f(\x)-f(\x')) \\
+\eta(\x') (1-\eta(\x)) \phi(f(\x')-f(\x))],
\end{multline}
and denote by the optimal expected $\phi$-risk $R^*_\phi=\inf_{f}R_\phi(f)$ where the infimum takes over all measurable functions. Given two instances $\x,\x'\in\X$, we denote by the conditional $\phi$-risk as
\begin{equation}\label{eq:con-risk}
C(\x,\x',\alpha)=\frac{1}{2p(1-p)}(\eta(\x)(1-\eta(\x'))\phi(\alpha) + \eta(\x')(1-\eta(\x))\phi(-\alpha)) ,
\end{equation}
where $\alpha=f(\x)-f(\x')$, and it holds that $R_\phi(f)=E_{\x,\x' \sim\mathcal{D}_\mathcal{X}^2} [C(\x, \x',\alpha)]$.
For convenience, we denote by $\eta=\eta(\x)$ and $\eta'=\eta(\x')$. Then, we define the optimal conditional $\phi$-risk
\begin{eqnarray}
H(\eta,\eta')&=&\inf_{\alpha\in\mathbb{R}}C(\x,\x',\alpha) \nonumber \\
&=&\frac{1}{2p(1-p)}\inf_{\alpha \in\mathbb{R}} \left\{\eta(1-\eta')\phi(\alpha) + \eta'(1-\eta) \phi(-\alpha)\right\},\label{eq:t0}
\end{eqnarray}
and further define
\begin{equation}\label{eq:t1}
H^-(\eta,\eta')=\frac{1}{2p(1-p)}\inf_{\alpha\colon \alpha(\eta-\eta')\leq 0} \left\{\eta(1-\eta') \phi(\alpha) + \eta'(1-\eta) \phi(-\alpha)\right\}.
\end{equation}

\section{AUC Consistency}\label{sec:mainres}
We first define the \emph{AUC consistency} as follows:
\begin{definition}
The surrogate loss $\phi$ is said to be consistent with AUC if for every sequence $\{f^{\langle n\rangle}(\x)\}_{n\geq1}$, the following holds over all distributions $\mathcal{D}$ on $\mathcal{X}\times\mathcal{Y}$:
\[
R_\phi(f^{\langle n\rangle})\to R_\phi^* \text{ then }R(f^{\langle n\rangle})\to R^*.
\]
\end{definition}

In binary classification, \cite{Bartlett:Jordan:McAuliffe2006} showed that the \emph{classification calibration} is sufficient and necessary to consistency of $0/1$ error. Motivated from this work, we generalize the \emph{calibration} to AUC as follows:
\begin{definition}
The surrogate loss $\phi$ is said to be calibrated if
\[
H^-(\eta,\eta')>H(\eta,\eta') \text{ for any }\eta\neq\eta'
\]
where $H$ and $H^-$ are defined by Eqns.~\eqref{eq:t0} and \eqref{eq:t1}, respectively.
\end{definition}
We will try to understand the relationship between calibration and AUC consistency. Recall that
\[
R^*_\phi=\inf_f{R_\phi(f)}= \inf_{f}E_{\x,\x'\sim\mathcal{D}^2_{\mathcal{X}}} C(\eta(\x),\eta(\x'),\alpha),
\]
and we first observe that
\begin{equation}\label{eq:inequality}
R^*_\phi=\inf_f{R_\phi(f)}\geq E_{\x,\x'\sim\mathcal{D}^2_{\mathcal{X}}} \inf_{\alpha}C(\eta(\x),\eta(\x'),\alpha).
\end{equation}
Notice that the equality in Eqn.~\eqref{eq:inequality} does not hold for many commonly-used surrogate losses such as hinge loss, least square hinge loss, least square loss, absolute loss, etc., which can be shown by the following lemma:
\begin{lemma}\label{lem2}
For hinge loss $\phi(t)=\max(0,1-t)$, least square hinge loss $\phi(t)= (\max(0,1-t))^2$, least square loss $\phi(t)=(1-t)^2$ and absolute loss $\phi(t)=|1-t|$, we have
\[
\inf_f{R_\phi(f)} > E_{\x,\x'\sim\mathcal{D}^2_{\mathcal{X}}}\inf_{\alpha}C(\eta(\x),\eta(\x'),\alpha).
\]
\end{lemma}

Lemma~\ref{lem2} shows that minimizing the expected $\phi$-risk $R_\phi(f)$ over the whole distribution is not equivalent to minimizing the conditional $\phi$-risk $C(\x,\x',\alpha)$ on each pair of instances from different class. Therefore, for pairwise surrogate loss, the study on AUC consistency should focus on the expected $\phi$-risk over the whole distribution rather than conditional $\phi$-risk on each pair of instances. This is quite different from binary classification where minimizing the expected risk over the whole distribution is equivalent to minimizing the conditional risk on each instance, and thus the study on consistency of binary classification focuses on the conditional risk as illustrated in \citep{Zhang2004ann,Bartlett:Jordan:McAuliffe2006}.

\proof We will present detailed proof for hinge loss by contradiction, and similar considerations could be made to other losses. Suppose that there exists a function $f$ such that
\[
{R_\phi(f)} = E_{\x,\x'\sim\mathcal{D}^2_{\mathcal{X}}} [\inf_{\alpha}C(\eta(\x),\eta(\x'),\alpha)].
\]
For simplicity, we consider three different instances $\x_1,\x_2,\x_3\in\mathcal{X}$ such that
\[
\eta(\x_1)<\eta(\x_2)<\eta(\x_3).
\]
The conditional risk of hinge loss is given by
\begin{eqnarray*}
C(\x,\x',\alpha)&=&\tfrac{1}{2p(1-p)}\eta(\x)(1-\eta(\x')) \max(0,1-\alpha)\\
&&+\tfrac{1}{2p(1-p)}\eta(\x')(1-\eta(\x))\max(0,1+\alpha),
\end{eqnarray*}
and minimizing $C(\x,\x',\alpha)$ gives $\alpha=-1$ if $\eta(\x)<\eta(\x')$. From the assumption that
\[
{R_\phi(f)} = E_{\x,\x'\sim\mathcal{D}^2_{\mathcal{X}}} \inf_{\alpha} C(\eta, \eta', \alpha),
\]
we have $f(\x_1)-f(\x_2)=-1$, $f(\x_1)-f(\x_3)=-1$ and $f(\x_2)-f(\x_3)=-1$; while they are contrary to each other. \qed

\subsection{Calibration is Necessary yet Insufficient for AUC Consistency}

We first prove that calibration is a necessary condition for AUC consistency by the following lemma:
\begin{lemma}\label{thm:cal2}
If the surrogate loss $\phi$ is consistent with AUC, then $\phi$ is calibrated, and for convex $\phi$, it is differentiable at $t=0$ with $\phi'(0)<0$.
\end{lemma}

The proof is partly motivated from \citep{Bartlett:Jordan:McAuliffe2006}, and we defer it to Section~\ref{sec:pf:cal2}. For the converse direction, we first observe that hinge loss $\phi(t)=\max(0,1-t)$ is inconsistent with respect to AUC as follows:
\begin{lemma}\label{lem:hinge}
For hinge loss $\phi(t)=\max(0,1-t)$, the surrogate loss $\Psi(f,\x,\x')= \phi(f(\x)- f(\x'))$ is inconsistent with AUC.
\end{lemma}

The detailed proof is deferred to Section~\ref{pf:lem:hinge}. In addition to hinge loss, the absolute loss $\phi(t)=|1-t|$ is also proven to be inconsistent with AUC:
\begin{lemma}\label{lem:abs}
For absolute loss $\phi(t)=|1-t|$, the surrogate loss $\Psi(f,\x,\x')= \phi(f(\x)- f(\x'))$ is inconsistent with AUC.
\end{lemma}

The detailed proof is presented in Section~\ref{pf:lem:abs}. It is noteworthy that hinge loss $\phi(t)=\max(0,1-t)$ and absolute loss $\phi(t)=|1-t|$ are convex with $\phi'(0)<0$, and thus they are calibrated, whereas Lemmas~\ref{lem:hinge} and \ref{lem:abs} show their inconsistency with AUC, respectively. Therefore, classification calibration is no longer a sufficient condition for AUC consistency.

Combining Lemmas~\ref{thm:cal2}-\ref{lem:abs}, we have
\begin{theorem}\label{thm:calt}
Calibration is necessary yet insufficient for AUC consistency.
\end{theorem}

This theorem shows that the study on AUC consistency is not parallel to that of binary classification where the classification calibration is necessary and sufficient for the consistency of $0/1$ error in \citep{Bartlett:Jordan:McAuliffe2006}. The main difference is that, for AUC consistency, minimizing the expected risk over the whole distribution is not equivalent to minimizing the conditional risk on each pair of instances as shown in Lemma~\ref{lem2}.

\subsection{Sufficient Condition for AUC Consistency}\label{sec:suf}
Based on the previous analysis, we present a new sufficient condition for AUC consistency, and the detailed proof is deferred to Section~\ref{sec:pf1}.
\begin{theorem}\label{thm:uniform-con}
The surrogate loss $\Psi(f,\x,\x')=\phi(f(\x)-f(\x'))$ is consistent with AUC if $\phi\colon \mathbb{R} \to\mathbb{R}$ is a convex, differentiable and non-increasing function with $\phi'(0)<0$.
\end{theorem}

\cite{Uematsu:Lee2011} proved the inconsistency of hinge loss and presented a sufficient condition, whereas our proof technique is considerably simpler than that of Uematsu and Lee (2011), especially for the proof of inconsistency of hinge loss. We will also provide a necessary condition in previous section and regret bounds later.

Based on Theorem~\ref{thm:uniform-con}, many surrogate losses are proven to be consistent with AUC as follows:
\begin{corollary}\label{coro:exp:con}
For exponential loss $\phi(t)=e^{-t}$, the surrogate loss $\Psi(f,\x,\x')= \phi(f(\x)- f(\x'))$ is consistent with AUC.
\end{corollary}
\begin{corollary}\label{coro:log:con}
For logistic loss $\phi(t)=\ln(1+e^{-t})$, the surrogate loss $\Psi(f,\x,\x')= \phi(f(\x)-f(\x'))$ is consistent with AUC.
\end{corollary}

\citet{Marron:Todd:Ahn2007} introduced the distance-weighted discrimination method to deal with the problems with high dimension yet small-size sample, and this method has been reformulated by \citet{Bartlett:Jordan:McAuliffe2006}, for any $\epsilon>0$, as follows:
\begin{equation}\label{eq:weight}
\phi(t)=\left\{
\begin{array}{l}
\frac{1}{t} \quad\quad\quad\quad\quad\quad\quad\text{ for }t\geq\epsilon,\\
\frac{1}{\epsilon}\left(2-\frac{t}{\epsilon}\right)\quad\quad\quad\text{ otherwise}.
\end{array}\right.
\end{equation}
Based on Theorem~\ref{thm:uniform-con}, we can also derive its consistency as follows:
\begin{corollary}
For distance-weighted loss $\phi$ given by Eqn.~\eqref{eq:weight} with $\epsilon>0$, the surrogate loss $\Psi(f,\x,\x')= \phi(f(\x)-f(\x'))$ is consistent with AUC.
\end{corollary}

It is noteworthy that the hinge loss $\phi(t)=\max(0,1-t)$ is not differentiable at $t=1$, and we cannot apply Theorem~\ref{thm:uniform-con} directly to study the consistency of hinge loss. Lemma~\ref{lem:hinge} proves its inconsistency and also shows the difficulty for consistency without differentiability, even if the surrogate loss function $\phi$ is convex and non-increasing with $\phi'(0)<0$. We now derive some variants of hinge loss that are consistent. For example, the \textit{$q$-norm hinge loss}:
\[
\phi(t)=(\max(0,1-t))^q ~~~\text{ for some }q>1.
\]
From Theorem~\ref{thm:uniform-con}, we can get the AUC consistency of the $q$-norm hinge loss:
\begin{corollary}
For $q$-norm hinge loss $\phi(t)=(\max(0,1-t))^q$ with $q>1$, the surrogate loss $\phi(f,\x,\x')=\phi(f(\x)-f(\x'))$ is consistent with AUC.
\end{corollary}
From this corollary, it is immediate to get the consistency for the \textit{least-square hinge loss} $\phi(t)= (\max(0,1-t))^2$. We further define the \textit{general hinge loss}, for any $\epsilon>0$, as:
\begin{equation}\label{eq:xxxloss}
\phi(t)=\left\{
\begin{array}{l}
1-t \quad\quad\quad\quad\quad\text{ for }t\leq1-\epsilon,\\
(t-1-\epsilon)^2/4\epsilon \ \ \text{ for }1-\epsilon\leq t<1+\epsilon,\\
0\quad\quad\quad\quad\quad\quad\ \,\text{ otherwise}.
\end{array}\right.
\end{equation}
It is easy to obtain the AUC consistency of general hinge loss from Theorem~\ref{thm:uniform-con}:
\begin{corollary}
For general hinge loss $\phi$ given by Eqn.~\eqref{eq:xxxloss} with $\epsilon>0$, the surrogate loss $\Psi(f,\x,\x')=\phi(f(\x)-f(\x'))$ is consistent with AUC.
\end{corollary}

Hinge loss is inconsistent with AUC, but we can use consistent surrogate loss, e.g., the general hinge loss, to approach hinge loss when $\epsilon\to0$. In addition, it is also interesting to derive other surrogate loss functions that are consistent with AUC under the guidance of Theorem~\ref{thm:uniform-con}.

\section{Regret Bounds}\label{sec:bounds}
In this section, we first present the regret bounds for exponential loss and logistic loss, and then study the regret bounds for general losses under the realizable setting.

\subsection{Regret Bounds for Exponential Loss and Logistic Loss}
Corollaries~\ref{coro:exp:con} and \ref{coro:log:con} show that the exponential loss and logistic loss are consistent with AUC, respectively. We further study their regret bounds based on the following special property:
\begin{lemma}\label{lem1}
For exponential loss and logistic loss, it holds that
\[
\inf_f{R_\phi(f)}=E_{\x,\x'\sim\mathcal{D}^2_{\mathcal{X}}} \inf_{\alpha} C(\eta(\x), \eta(\x'), \alpha).
\]
\end{lemma}
\proof We provide the detailed proof for exponential loss, and similar consideration could be made to logistic loss. Fixing an instance $\x_0\in\mathcal{X}$ and $f(\x_0)$, we set
\[
f(\x)=f(\x_0)+\frac{1}{2}\ln\frac{\eta(\x)(1-\eta(\x_0))}{\eta(\x_0)(1-\eta(\x))} ~~\text{ for }\x\neq \x_0.
\]
It remains to prove $R(f)=E_{\x,\x'\sim\mathcal{D}^2_{\mathcal{X}}} \inf_{\alpha}C(\eta(\x),\eta(\x'),\alpha)$. Based on the above equation, we have, for instances $\x_1,\x_2\in\mathcal{X}$:
\[
f(\x_1)-f(\x_2)=\frac{1}{2}\ln\frac{\eta(\x_1)(1-\eta(\x_2))}{\eta(\x_2)(1-\eta(\x_1))},
\]
which exactly minimizes $C(\eta(\x_1),\eta(\x_2),\alpha)$ when $\alpha= f(\x_1)-f(\x_2)$.\qed \vspace{+2mm}

It is noteworthy that Lemma~\ref{lem1} is specific to the exponential loss and logistic loss, and it does not hold for other surrogate loss functions such as hinge loss, general hinge loss, $q$-norm hinge loss, etc. Based on Lemma~\ref{lem1}, we study the regret bounds for exponential loss and logistic loss by focusing on conditional risk. We first present a general theorem as follows:
\begin{theorem}\label{thm:con}
For some $\kappa_0>0$ and $0<\kappa_1\leq 1$, we have
\[
R(f)-R^*\leq \kappa_0(R_\phi(f)-R_\phi^*)^{\kappa_1},
\]
if the surrogate loss $\phi$ satisfies $\inf_{f}R_\phi(f)=E_{\x,\x'\sim\mathcal{D}^2_{\mathcal{X}}} \inf_{\alpha} C[\eta(\x), \eta(\x'),\alpha]$, and if $f^*\in\mathop{\arg\inf}_{f}R_\phi(f)$ is such that
\begin{eqnarray*}
&(f^*(\x)-f^*(\x'))(\eta(\x)-\eta(\x'))>0 \text{ for } \eta(\x)\neq\eta(\x'), \text{ and}&\\
&\tfrac{|\eta(\x)-\eta(\x')|}{2p(1-p)}\leq \kappa_0 \big(C(\eta(\x),\eta(\x'),0) -C(\eta(\x),\eta(\x'),f^*(\x)-f^*(\x'))\big)^{\kappa_1}.&
\end{eqnarray*}
\end{theorem}

This proof is partly motivated from \citet{Zhang2004ann} and we defer it to Section~\ref{sec:pf2}. Based on this theorem, we can get the following regret bounds for the exponential loss and logistic loss:
\begin{corollary}\label{coro:exp}
For exponential loss, it holds that $R(f)-R^*\leq \sqrt{R_\phi(f)-R_\phi^*}$\,.
\end{corollary}
\begin{corollary}\label{coro:logistic}
For logistic loss, it holds that $R(f)-R^*\leq 2\sqrt{R_\phi(f)-R_\phi^*}$\,.
\end{corollary}

The detailed proofs of Corollaries~\ref{coro:exp} and \ref{coro:logistic} are given in Section~\ref{sec:pf:app1} and \ref{sec:pf:app2}, respectively.

\subsection{Regret Bounds for Realizable Setting}
Now we define the realizable setting as:
\begin{definition}
A distribution $\mathcal{D}$ is said to be realizable if $\eta(\x)(1-\eta(\x))=0$ for each $\x\in\mathcal{X}$.
\end{definition}

Such setting have been studied for bipartite ranking \citep{Rudin:Schapire2009} and multi-class classification \citep{Long:Servedio2013}. Under this setting, we have the regret bounds as follows:
\begin{theorem}
For some $\kappa>0$, we have
\[
R(f)-R^*\leq \kappa(R_\phi(f)-R_\phi^*),
\]
if $R^*_\phi=0$, and if $\phi(t)\geq1/\kappa$ for $t\leq0$ and $\phi(t)\geq0$ for $t>0$.
\end{theorem}
\proof For convenience, denote by $\mathcal{D}_+$ and $\mathcal{D}_-$ the positive and negative instance distributions, respectively. From Eqn.~\eqref{eq:truerisk}, we have
\[
R(f)=\tfrac{1}{2p(1-p)}E_{\x\sim\mathcal{D}_+,\x'\sim\mathcal{D}_-}[I[f(\x)<f(\x')] +I[f(\x)=f(\x')]/2],
\]
and thus $R^*=\inf_{f}[R(f)]=0$ when $f(\x)>f(\x')$. From Eqn.~\eqref{eq:phi-risk}, we get the $\phi$-risk $R_\phi(f) =E_{\x\sim\mathcal{D}_+,\x'\sim\mathcal{D}_-} [\phi(f(\x)-f(\x'))]$. Then
\begin{eqnarray*}
R(f)-R^*&=&\tfrac{1}{2p(1-p)}E_{\x\sim\mathcal{D}_+,\x'\sim\mathcal{D}_-}[I[f(\x)<f(\x')] +I[f(\x)=f(\x')]/2]\\
&\leq&\tfrac{1}{2p(1-p)} E_{\x\sim\mathcal{D}_+,\x'\sim\mathcal{D}_-}[\kappa\phi(f(\x)-f(\x'))] =\kappa(R_\phi(f)-R^*_\phi),
\end{eqnarray*}
which completes the proof.\qed \vspace{+2mm}

Based on this theorem, we have the following regret bounds:
\begin{corollary}
For exponential loss, hinge loss, general hinge loss, $q$-norm hinge loss, and least square loss, we have
\[
R(f)-R^*\leq R_\phi(f)-R_\phi^*,
\]
and for logistic loss, we have
\[
R(f)-R^*\leq \tfrac{1}{\ln2}(R_\phi(f)-R_\phi^*).
\]
\end{corollary}
It is noteworthy that the hinge loss is consistent with AUC under the realizable setting yet inconsistent for the general case as shown in Lemma~\ref{lem:hinge}. Corollaries~\ref{coro:exp} and \ref{coro:logistic} show regret bounds for exponential loss and logistic loss in the general case, respectively, whereas the above corollary provides tighter regret bounds under the realizable setting.

\section{Equivalence Between AUC and Accuracy Optimization with Exponential Loss}\label{sec:rel}

In this section, we analyze the relationship of exponential loss for AUC and accuracy, and present regret bounds to show their equivalence.

In binary classification, we always learn a score function $f\in\X\to\mathbb{R}$, and make predictions based on $\sgn[f(\x)]$. The goal is to improve the accuracy by minimizing
\begin{eqnarray*}
R_\text{acc}(f)&=&E_{\left(\x,y\right)\sim\mathcal{D}} \left[I\left[yf(\x)<0\right]\right]\\
&=&E_{\x}\left[\eta\left(\x\right)I\left[f\left(\x\right)<0\right]+ \left(1-\eta\left(\x\right)\right)I\left[f\left(\x\right)>0\right]\right].
\end{eqnarray*}
We denote by $R_\text{acc}^*=\inf_{f}R_\text{acc}(f)$ where the infimum takes over all measurable functions, and it is easy to obtain the set of optimal solutions for accuracy as follows:
\[
\mathcal{B}_{\text{acc}}=\{f\colon R_\text{acc}(f)=R_\text{acc}^*\}=\{f\colon f(\x)(\eta(\x)-1/2)>0 \text{ for }\eta(\x)\neq1/2\}.
\]
In binary classification, the most popular formulation for surrogate losses is given by:
\[
\phi_\text{acc}(f(\x),y)=\phi(yf(\x)),
\]
where $\phi$ is a convex function, e.g., hinge loss $\phi(t)=\max(0, 1-t)$ \citep{vapnik98}, exponential loss $\phi(t)=e^{-t}$ \citep{Freund:Schapire1997}, logistic loss $\phi(t)=\ln(1+e^{-t})$ \citep{Friedman:Hastie:Tibshirani2000}, etc. We define $\phi_\text{acc}$-risk as
\[
R_{\phi_\text{acc}}(f)=E_{(\x,y)\sim\mathcal{D}}[\phi(yf(\x))]= E_{\x}[C_\text{acc}(\eta(\x),f(\x))]
\]
where $C_\text{acc}(\eta(\x),f(\x))= \eta(\x)\phi(f(\x))+ (1-\eta(\x)) \phi(-f(\x))$. Further, we denote by $R_{\phi_\text{acc}}^*=\inf_{f}R_{\phi_\text{acc}}(f)$, where the infimum takes over all measurable functions.

We begin with a regret bound as follows:
\begin{theorem}\label{thm:acc:auc}
For a classifier $f$ and exponential loss $\phi(t)=e^{-t}$, we have
\[
p(1-p)(R_\phi(f)-R_\phi^*)\leq R_{\phi_\text{acc}}(f) (R_{\phi_\text{acc}}(f)-R_{\phi_\text{acc}}^*).
\]
\end{theorem}
The detailed proof is presented in Section~\ref{pf:thm:acc:auc}. This theorem shows that a good classifier, which is learned by optimizing the exponential loss of accuracy, optimizes the pairwise exponential loss of AUC.

For a ranking function $f$, we will first find some proper threshold to construct classifier. Here, we present a simple way to select a threshold by
\[
t_f^*\in\mathop{\arg\min}_{t\in(-\infty,+\infty)} R_{\phi_\text{acc}}(f-t)= \mathop{\arg\min}_{t\in(-\infty,+\infty)}E_{\x}\big[\eta(\x)e^{-f(\x)+t} +(1-\eta(\x))e^{f(\x)-t}\big],
\]
and it is easy to get, for convex and smooth exponential loss, that
\[
t_f^*=\tfrac{1}{2}\ln E_\x[\eta(\x)e^{-f(\x)}]- \tfrac{1}{2}\ln E_\x[(1-\eta(\x))e^{f(\x)}].
\]
Based on such threshold, we have
\begin{theorem}\label{thm:auc:acc}
For a score ranking function $f$ and exponential loss $\phi(t)=e^{-t}$, we have
\[
R_{\phi_\text{acc}}(f-t^*_f)-R_{\phi_\text{acc}}^*\leq 2\sqrt{p(1-p)(R_{\phi}(f)-R^*_{\phi})}
\]
by selecting the threshold $t_f^*=\tfrac{1}{2}\ln E_\x[\eta(\x)e^{-f(\x)}]- \tfrac{1}{2}\ln E_\x[(1-\eta(\x))e^{f(\x)}]$.
\end{theorem}

The proof is presented in Section~\ref{pf:thm:auc:acc}. From this theorem, we can see that a score ranking function $f(\x)$, which is learned by optimizing the pairwise exponential loss of AUC, optimizes the exponential loss of accuracy by selecting a proper threshold.

Together with Corollary~\ref{coro:exp}, Theorems~\ref{thm:acc:auc} and \ref{thm:auc:acc}, and \cite[Theorem 2.1]{Zhang2004ann}, we have
\begin{theorem}
For a classifier $f(\x)$ and exponential loss $\phi(t)=e^{-t}$, we have
\begin{eqnarray*}
&R(f)-R^*\leq \left(\tfrac{R_{\phi_\text{acc}}(f)}{p(1-p)}(R_{\phi_\text{acc}}(f)-R_{\phi_\text{acc}}^*) \right)^{1/2}&\\
&R_\text{acc}(f)-R_\text{acc}^*\leq \sqrt{2}(R_{\phi_\text{acc}}(f)-R_{\phi_\text{acc}}^*)^{1/2}\ .&
\end{eqnarray*}
For a ranking function $f(\x)$ and exponential loss $\phi(t)=e^{-t}$, we have
\begin{eqnarray*}
&R(f)-R^*\leq (R_\phi(f)-R_\phi^*)^{1/2}&\\
&R_\text{acc}(f-t^*_f)-R_\text{acc}^*\leq 2(p(1-p)(R_{\phi}(f)-R^*_{\phi}))^{1/4}\ .&
\end{eqnarray*}
by selecting the threshold $t_f^*=\tfrac{1}{2}\ln E_\x[\eta(\x)e^{-f(\x)}]- \tfrac{1}{2}\ln E_\x[(1-\eta(\x))e^{f(\x)}]$.
\end{theorem}

This theorem shows the asymptotic equivalence between the exponential surrogate loss of accuracy and the pairwise exponential surrogate loss of AUC. Thus, the surrogate loss $\phi_\text{acc}(f({\bm x}),y) = e^{-yf({\bm x})}$ of accuracy is consistent with AUC, while the pairwise surrogate loss $\phi(f,{\bm x},{\bm x}')=e^{-(f({\bm x})-f({\bm x}'))}$ of AUC is consistent with accuracy by choosing a proper threshold. One direct consequence of this theorem is: AdaBoost and RankBoost are equivalent asymptotically, i.e., both of them optimize AUC and accuracy simultaneously in infinite sample, because AdaBoost and RankBoost essentially optimize the surrogate loss $\phi_\text{acc}(f({\bm x}),y) = e^{-yf({\bm x})}$ and $\phi(f,{\bm x},{\bm x}')=e^{-(f({\bm x})-f({\bm x}'))}$, respectively.

\cite{Rudin:Schapire2009} has established the equivalence between AdaBoost
and RankBoost for finite training sample. For that purpose, they assumed that the negative and positive classes contributed equally, although this is often not the fact in practice. Our work does not make such assumption, and we consider the limit of infinite sample. Moreover, our regret bounds, which shows the equivalence between AUC and accuracy optimization with exponential surrogate loss, provides a new explanation to the equivalence between AdaBoost and RankBoost.

\section{Proofs}\label{sec:pf}
In this section, we provide some detailed proofs for our results.

\subsection{Proof of Lemma~\ref{thm:cal2}}\label{sec:pf:cal2}

If $\phi$ is not calibrated, then there exist $\eta_0$ and $\eta_0'$ s.t. $\eta_0>\eta_0'$ and $H^-(\eta_0,\eta_0')=H(\eta_0,\eta_0')$, that is,
\begin{eqnarray*}
&&\tfrac{1}{2p(1-p)}\inf_{\alpha \in\mathbb{R}} \left\{\eta_0(1-\eta_0')\phi(\alpha) + \eta_0'(1-\eta_0) \phi(-\alpha)\right\}\\
&&=\tfrac{1}{2p(1-p)}\inf_{\alpha\colon \alpha(\eta_0-\eta_0')\leq 0} \left\{\eta_0(1-\eta_0') \phi(\alpha) + \eta_0'(1-\eta_0) \phi(-\alpha)\right\}.
\end{eqnarray*}
This implies the existence of some $\alpha_0\leq 0$ such that
\[
\eta_0(1-\eta_0') \phi(\alpha_0) + \eta_0'(1-\eta_0) \phi(-\alpha_0)= \inf_{\alpha \in\mathbb{R}} \left\{\eta_0(1-\eta_0')\phi(\alpha) + \eta_0'(1-\eta_0) \phi(-\alpha) \right\}.
\]
We consider an instance space $\mathcal{X}=\{\x_1,\x_2\}$ with marginal probability $\Pr[\x_i]=1/2$ and conditional probability $\eta(\x_1)=\eta_0$ and $\eta(\x_2)=\eta_0'$. We then construct a sequence $\{f^{\langle n\rangle}\}_{n\neq1}$ by picking up $f^{\langle n\rangle}(\x_1)=f^{\langle n\rangle}(\x_2)+\alpha_0$, and it is easy to get that
\[
R_\phi(f^{\langle n\rangle})\to R_\phi^*\text{ yet }R(f^{\langle n\rangle})-R^* = (\eta_0-\eta_0')/8\text{ as }n\to\infty.
\]
This shows the inconsistency of $\phi$; therefore, calibration is a necessary condition for AUC consistency.

For convex $\phi$, we will show that the condition that $\phi$ is differentiable at $t=0$ and $\phi'(0)<0$ is necessary for AUC consistency. For convenience, we consider the instance space $\mathcal{X}=\{\x_1,\x_2\}$ with marginal probability $\Pr[\x_1]=\Pr[\x_2]=1/2$ and conditional probability $\eta(\x_1)=\eta_1$ and $\eta(\x_2)=\eta_2$.

We first prove that if the consistent surrogate loss $\phi$ is differentiable at $t=0$, then $\phi'(0)<0$. Assume $\phi'(0)\geq0$, and for convex $\phi$, we have
\begin{multline*}
\eta_1(1-\eta_2)\phi(\alpha)+ \eta_2(1-\eta_1)\phi(-\alpha) \geq(\eta_1-\eta_2)\alpha\phi'(0)\\
+(\eta_1(1-\eta_2)+\eta_2(1-\eta_1))\phi(0) \geq(\eta_1(1-\eta_2)+\eta_2(1-\eta_1))\phi(0)
\end{multline*}
for $(\eta_1-\eta_2)\alpha\geq0$. This follows that
\begin{eqnarray}
\lefteqn{2p(1-p)H(\eta_1,\eta_2)=\inf_{\alpha\in\mathbb{R}}\left\{\eta_1(1-\eta_2)\phi(\alpha)+ \eta_2(1-\eta_1)\phi(-\alpha)\right\}\nonumber}\\
&=&\min\{\left\{\eta_1(1-\eta_2)\phi(0)+ \eta_2(1-\eta_1)\phi(0)\right\}, \nonumber\\
&&~~~~~~~~\left.\inf_{(\eta_1-\eta_2)\alpha\leq0}\left\{\eta_1(1-\eta_2)\phi(\alpha)+ \eta_2(1-\eta_1)\phi(-\alpha)\right\}\right\}\nonumber\\
&=&\inf_{(\eta_1-\eta_2)\alpha\leq0}\left\{\eta_1(1-\eta_2)\phi(\alpha)+ \eta_2(1-\eta_1)\phi(-\alpha)\right\}\nonumber\\
&=&2p(1-p)H^-(\eta_1,\eta_2),\label{eq:tmp4}
\end{eqnarray}
which implies that $\phi$ is not calibrated, and it is contrary to consistency of $\phi$.

We now prove that convex loss $\phi$ is differentiable at $t=0$. Assume that $\phi$ is not differentiable at $t=0$. We can find subgradients $g_1>g_2$ such that
\[
\phi(t)\geq g_1t+\phi(0) \text{ and } \phi(t)\geq g_2 t+\phi(0) \text{ for }t\in\mathbb{R},
\]
and it is sufficient to consider the following cases:
\begin{enumerate}
\item For $g_1>g_2\geq0$, we select $\eta_1=g_1/(g_1+g_2)$ and $\eta_2 =g_2/(g_1+g_2)$. It is obvious that $\eta_1>\eta_2$, and for any $\alpha\geq0$, we have
    \begin{eqnarray*}
    \lefteqn{\eta_1(1-\eta_2)\phi(\alpha)+ \eta_2(1-\eta_1)\phi(-\alpha)}\\
    &&\geq\eta_1(1-\eta_2)(g_2\alpha+\phi(0))+ \eta_2(1-\eta_1) (-g_1\alpha+\phi(0))\\
    &&=(g_1-g_2)\eta_1\eta_2\alpha+ (\eta_1(1-\eta_2) +\eta_2(1-\eta_1))\phi(0)\\
    &&\geq(\eta_1(1-\eta_2) +\eta_2(1-\eta_1))\phi(0);
    \end{eqnarray*}
\item For $g_1\geq0>g_2$ or $g_1>0\geq g_2$, we select $\eta_1=1$ and $\eta_2=1/2$, and for any $\alpha\geq0$, it holds that
    \begin{eqnarray*}
      \lefteqn{\eta_1(1-\eta_2)\phi(\alpha)+ \eta_2(1-\eta_1)\phi(-\alpha)}\\
      &&\geq\eta_1(1-\eta_2)(g_1\alpha+\phi(0))+ \eta_2(1-\eta_1) (-g_2\alpha+\phi(0))\\
      &&=g_1\alpha/2+(\eta_1(1-\eta_2)+\eta_2(1-\eta_1))\phi(0)\\
      &&\geq(\eta_1(1-\eta_2) +\eta_2(1-\eta_1))\phi(0);
      \end{eqnarray*}
\item For $0\geq g_1>g_2$, we select $\eta_1=(|g_1|+|g_1-g_2|/2) /(|g_1+g_2|)$ and $\eta_2=|g_1|/(|g_1+g_2|)$. We have $\eta_1>\eta_2$, and for any $\alpha\geq0$, it holds that
    \begin{eqnarray*}
      \lefteqn{\eta_1(1-\eta_2)\phi(\alpha)+ \eta_2(1-\eta_1)\phi(-\alpha)}\\
      &&\geq\eta_1(1-\eta_2)(g_1\alpha+\phi(0))+ \eta_2(1-\eta_1) (-g_2\alpha+\phi(0))\\
      &&=(\eta_1(1-\eta_2) +\eta_2(1-\eta_1))\phi(0).
      \end{eqnarray*}
\end{enumerate}
Therefore, for any $g_1$ and $g_2$, there exist $\eta_1$ and $\eta_2$ such that
\[
\eta_1(1-\eta_2)\phi(\alpha)+ \eta_2(1-\eta_1)\phi(-\alpha) \geq (\eta_1(1-\eta_2) +\eta_2(1-\eta_1))\phi(0)
\]
for $(\eta_1-\eta_2)\alpha\geq0$. Similarly to Eqn.~\eqref{eq:tmp4}, we have $H(\eta_1,\eta_2)=H^-(\eta_1,\eta_2)$, which is contrary to the consistency of $\phi$. \qed

\subsection{Proof of Lemma~\ref{lem:hinge}}\label{pf:lem:hinge}

For simplicity, we consider a special instance space $\X=\{\x_1, \x_2, \x_3\}$. For $1\leq i\leq 3$, we assume that the marginal probability $\Pr[\x_i]=1/3$ and conditional probability $\eta_i=\eta(\x_i)$ satisfy
\[
\eta_1<\eta_2<\eta_3, 2\eta_2<\eta_1+\eta_3,\text{ and } 2\eta_1>\eta_2+\eta_1\eta_3.
\]
We further write $f_i=f(\x_i)$ for $1\leq i\leq 3$, and Eqn.~\eqref{eq:phi-risk} gives
\begin{eqnarray*}
\lefteqn{R_\phi(f)=\kappa_0}\\
&&+\kappa_1\{\eta_1(1-\eta_2)\max(0,1+f_2-f_1)+\eta_2(1-\eta_1)\max(0,1+f_1-f_2)\}\\
&&+\kappa_1\{\eta_1(1-\eta_3)\max(0,1+f_3-f_1)+\eta_3(1-\eta_1)\max(0,1+f_1-f_3)\}\\
&&+\kappa_1\{\eta_2(1-\eta_3)\max(0,1+f_3-f_2)+\eta_3(1-\eta_2)\max(0,1+f_2-f_3)\},
\end{eqnarray*}
where $\kappa_0>0$ and $\kappa_1>0$ are constants and independent to $f$. Minimizing $R_\phi(f)$ yields the optimal expected $\phi$-risk
\[
R_\phi^*=\kappa_0+\kappa_1(3\eta_1+3\eta_2-2\eta_1\eta_2-2\eta_1\eta_3-2\eta_2\eta_3)
\]
when $f^*=(f^*_1,f^*_2,f^*_3)$ s.t. $f^*_1=f^*_2=f^*_3-1$. Note that $f'=(f'_1, f'_2, f'_3)$ s.t. $f'_1+1=f'_2=f'_3-1$ is not the optimal solution w.r.t. hinge loss since
\begin{eqnarray*}
R_\phi(f')&=&\kappa_0+\kappa_1(5\eta_1+2\eta_2-2\eta_1\eta_2-3\eta_1\eta_3-2\eta_2\eta_3)\\
&=&R_\phi^*+\kappa_1(2\eta_1-\eta_2-\eta_1\eta_3)>R_\phi^*
\end{eqnarray*}
where we use $2\eta_1>\eta_2+\eta_1\eta_3$.

We now construct a sequence $\{f^{\langle n\rangle}\}_{ n\geq1}$ by choosing $f^{\langle 1\rangle}(\x_1)=f^{\langle 1\rangle}(\x_2)=f^{\langle 1\rangle}(\x_3)-1$ and $f^{\langle n\rangle}(\x)=f^{\langle 1\rangle}(\x)$ for $n>1$. Then, it holds that
\[
R_\phi(f^{\langle n\rangle})= R_\phi^* \text{ yet } R(f^{\langle n\rangle})-R^*=\kappa_1(\eta_2-\eta_1)/2 \text{ for } n\geq1.
\]
Therefore, there exists a sequence $\{f^{\langle n\rangle}\}_{n\geq1}$ such that $R_\phi(f^{\langle n\rangle}) \rightarrow R_\phi^*$ yet $R(f^{\langle n\rangle})\nrightarrow R^*$, and this completes the proof.\qed

\subsection{Proof of Lemma~\ref{lem:abs}}\label{pf:lem:abs}

Similarly to the proof of Lemma~\ref{lem:hinge}, we consider the instance space $\mathcal{X}=\{\x_1,\x_2,\x_3\}$ with marginal probability $\Pr[\x_i]=1/3$ and conditional probability $\eta_i=\eta(\x_i)$ such that
\[
\eta_1<\eta_2<\eta_3 \text{ and }  2\eta_2>\eta_1+\eta_3.
\]
We write $f_i=f(\x_i)$, and Eqn.~\eqref{eq:phi-risk} gives
\begin{eqnarray*}
\lefteqn{R_\phi(f)=\kappa_0+\kappa_1 \big(\eta_1(1-\eta_2)|1+f_2-f_1|+\eta_2(1-\eta_1)|1+f_1-f_2|}\\
&&\quad\quad\quad\quad\quad\quad+\eta_1(1-\eta_3) |1+f_3-f_1|+\eta_3(1-\eta_1)|1+f_1-f_3|\\
&&\quad\quad\quad\quad\quad\quad+\eta_2(1-\eta_3) |1+f_3-f_2|+\eta_3(1-\eta_2)|1+f_2-f_3|\big),
\end{eqnarray*}
where $\kappa_0>0$ and $\kappa_1>0$ are constants and independent to $f$. Minimizing $R_\phi(f)$ gives
\[
R_\phi^*=\kappa_0+\kappa_1(4\eta_1+\eta_2+ \eta_3-2\eta_1\eta_2-2\eta_1\eta_3-2\eta_2\eta_3)
\]
when $f^*=(f^*_1,f^*_2,f^*_3)$ s.t. $f^*_1=f^*_2-1=f^*_3-1$. Also, note that $f'=(f'_1, f'_2, f'_3)$ s.t. $f'_1+1=f'_2=f'_3-1$ is not a optimal solution w.r.t. absolute loss since
\begin{eqnarray*}
R_\phi(f')&=&\kappa_0+\kappa_1(5\eta_1+2\eta_2+\eta_3-2\eta_1\eta_2-4\eta_1\eta_3-2\eta_2\eta_3)\\
&=&R_\phi^*+\kappa_1(\eta_1+\eta_2-2\eta_1\eta_3)>R_\phi^*
\end{eqnarray*}
where $\eta_1+\eta_2-2\eta_1\eta_3\geq\eta_2-\eta_1\eta_3>(\eta_1+\eta_3)/2 -\eta_1\eta_3\geq0$.

We can construct a sequence $\{f^{\langle n\rangle}\}_{ n\geq1}$ by choosing $f^{\langle 1\rangle}(\x_1)=f^{\langle 1\rangle}(\x_2)-1=f^{\langle 1\rangle}(\x_3)-1$ and $f^{\langle n\rangle}(\x)=f^{\langle 1\rangle}(\x)$ for $n>1$. Then, it holds that
\[
R_\phi(f^{\langle n\rangle})= R_\phi^* \text{ yet } R(f^{\langle n\rangle})-R^*=\kappa_1(\eta_3-\eta_2)/2 \text{ for } n\geq1.
\]
Therefore, there exists a sequence $\{f^{\langle n\rangle}\}_{n\geq1}$ such that $R_\phi(f^{\langle n\rangle}) \rightarrow R_\phi^*$ yet $R(f^{\langle n\rangle})\nrightarrow R^*$, and this completes the proof. \qed

\subsection{Proof of Theorem~\ref{thm:uniform-con}}\label{sec:pf1}

We begin with the following lemma, which is crucial to the proof of Theorem~\ref{thm:uniform-con}.
\begin{lemma}\label{thm:uniform-tmp}
For surrogate loss $\phi(f,x,x')=\phi(f(x)-f(x'))$, it holds that
\[
\inf_{f\notin \mathcal{B}}R_\phi(f)>\inf_{f}R_\phi(f)
\]
if $\phi\colon \mathbb{R} \to\mathbb{R}$ is a convex, differentiable and non-increasing function with $\phi'(0)<0$.
\end{lemma}
\proof From the $\phi$-risk's definition in Eqn.~\eqref{eq:phi-risk}, we have
\begin{multline*}
R_\phi(f)=\frac{1}{2p(1-p)}\int_{\X}\int_\X \eta(\x)(1-\eta(\x'))\phi(f(\x)-f(\x'))+\\
\eta(\x')(1-\eta(\x))\phi(f(\x')-f(\x)) d\Pr(\x)d\Pr(\x')
\end{multline*}
We proceed by contradiction, and suppose that
\[
\inf\nolimits_{f\notin \mathcal{B}}R_\phi(f)=\inf\nolimits_{f}R_\phi(f).
\]
This implies that there exists an optimal function $f^*$ such that $R_\phi(f^*)=\inf_{f}R_\phi(f)$ and $f^*\notin \mathcal{B}$, i.e., for some $\x_1,\x_2\in\mathcal{X}$, it holds that $f^*(\x_1) \leq f^*(\x_2)$ yet $\eta(\x_1)>\eta(\x_2)$.

We introduce a function $h_1$ s.t. $h_1(\x)=0$ if $\x\neq\x_1$ and $h_1(\x_1)=1$ otherwise. Also, we write $g(\gamma)=R_\phi(f^*+\gamma h_1)$ for any $\gamma\in\mathbb{R}$, and thus $g$ is convex since $\phi$ is convex. For optimal function $f^*$, we have $g'(0)=0$ which implies that
\begin{multline}\label{eq:temp1}
\int_{\X\setminus \x_1}\eta(\x_1)(1-\eta(\x))\phi'(f^*(\x_1)-f^*(\x)) \\ -\eta(\x)(1-\eta(\x_1))\phi'(f^*(\x)-f^*(\x_1))d\Pr(\x)=0.
\end{multline}
In a similar manner, we could introduce another function $h_2$ s.t. $h_2(\x)=0$ if $\x\neq\x_2$ and $h_2(\x_2)=1$ otherwise, and further derive
\begin{multline}\label{eq:temp2}
\int_{\X\setminus \x_2}\eta(\x_2)(1-\eta(\x))\phi'(f^*(\x_2)-f^*(\x)) \\ -\eta(\x)(1-\eta(\x_2))\phi'(f^*(\x)-f^*(\x_2))d\Pr(\x)=0.
\end{multline}
Combining Eqns.~\eqref{eq:temp1} and \eqref{eq:temp2} gives
\begin{multline}\label{eq:tmp1}
\int\limits_{\X\setminus\{\x_1,\x_2\}}\eta(\x)\big((1-\eta(\x_2))\phi'(f^*(\x)-f^*(\x_2))- (1-\eta(\x_1))\phi'(f^*(\x)-f^*(\x_1))\big)\\
+(1-\eta(\x))\big(\eta(\x_1)\phi'(f^*(\x_1)-f^*(\x))- \eta(\x_2)\phi'(f^*(\x_2)-f^*(\x))\big)d\Pr(\x)\\
+(\Pr(\x_1)+\Pr(\x_2))\big(\eta(\x_1)(1-\eta(\x_2))\phi'(f^*(\x_1)-f^*(\x_2)) \\ -\eta(\x_2)(1-\eta(\x_1))\phi'(f^*(\x_2)-f^*(\x_1))\big)=0.
\end{multline}
For convex differentiable and non-increasing function $\phi$, we have $\phi'(t_1)\leq\phi'(t_2)\leq0$ if $t_1\leq t_2$; therefore,  $\phi'(f^*(\x_1)-f^*(\x))\leq \phi'(f^*(\x_2)-f^*(\x))\leq0$ if $f^*(\x_1)\leq f^*(\x_2)$. This follows
\begin{equation}\label{eq:tmp2}
\eta(\x_1)\phi'(f^*(\x_1)-f^*(\x))- \eta(\x_2)\phi'(f^*(\x_2)-f^*(\x))\leq0
\end{equation}
for $\eta(\x_1)>\eta(\x_2)$. In a similar manner, we have
\begin{equation}\label{eq:tmp3}
(1-\eta(\x_2))\phi'(f^*(\x)-f^*(\x_2))- (1-\eta(\x_1))\phi'(f^*(\x)-f^*(\x_1))\leq0.
\end{equation}

If $f^*(\x_1)=f^*(\x_2)$, then we have
\begin{multline*}
\eta(\x_1)(1-\eta(\x_2))\phi'(f^*(\x_1)-f^*(\x_2))\\
-\eta(\x_2)(1-\eta(\x_1))\phi'(f^*(\x_2)-f^*(\x_1))=(\eta(\x_1)-\eta(\x_2))\phi'(0)<0
\end{multline*}
from $\phi'(0)<0$ and $\eta(\x_1)>\eta(\x_2)$, which is contrary to Eqn.~\eqref{eq:tmp1} by combining Eqns.~\eqref{eq:tmp2} and \eqref{eq:tmp3}.

If $f^*(\x_1)<f^*(\x_2)$, then we have $\phi'(f^*(\x_1)-f^*(\x_2))\leq\phi'(0)<0$ and $\phi'(f^*(\x_1)-f^*(\x_2))\leq \phi'(f^*(\x_2)-f^*(\x_1))\leq0$. This follows
\[
\eta(\x_1)(1-\eta(\x_2))\phi'(f^*(\x_1)-f^*(\x_2)) -\eta(\x_2)(1-\eta(\x_1))\phi'(f^*(\x_2)-f^*(\x_1))<0
\]
which is also contrary to Eqn.~\eqref{eq:tmp1} by combining Eqns.~\eqref{eq:tmp2} and \eqref{eq:tmp3}. Hence, this lemma follows as desired.\qed\vspace{0.15in}

\noindent\textbf{Proof of Theorem~\ref{thm:uniform-con}.} From Lemma~\ref{thm:uniform-tmp}, we set
\[
\delta=\inf_{f\notin \mathcal{B}}R_\phi(f)-\inf_{f}R_\phi(f)>0.
\]
Let $\{f^{\langle n\rangle}\}_{n\geq0}$ be an any sequence such that $R_\phi( f^{\langle n \rangle} )\to R^*_\phi$. Then, there exists an integer $N_0>0$ such that
\[
R_\phi(f^{\langle n \rangle})-R_\phi^*<\delta/2 \text{ for }n\geq N_0.
\]
This immediately yields that $f^{\langle n \rangle}\in\mathcal{B}$ for $n\geq N_0$ from the contrary that
\[
R_\phi(f)-R^*_\phi=R_\phi(f)-\inf_{f'\notin\mathcal{B}}R_\phi(f')+ \inf_{f'\notin\mathcal{B}}R_\phi(f')-R^*_\phi>\delta \text{ if }f\notin\mathcal{B}.
\]
Therefore, we have $R(f^{\langle n \rangle})=R^*$ for $n\geq N_0$, which completes the proof.\qed

\subsection{Proof of Theorem~\ref{thm:con}}\label{sec:pf2}

From Eqns.~\eqref{eq:truerisk} and \eqref{eq:AUCBayset}, we have
\begin{eqnarray*}
\lefteqn{2p(1-p)(R(f)-R^*)}\\
&=&E_{\eta(\x)>\eta(\x'),f(\x)<f(\x')}[\eta(\x)-\eta(\x')] +\tfrac{1}{2}E_{\eta(\x)>\eta(\x'),f(\x)=f(\x')}[\eta(\x)-\eta(\x')] \\
 & &+E_{\eta(\x)<\eta(\x'),f(\x)>f(\x')}[\eta(\x')-\eta(\x)] +\tfrac{1}{2}E_{\eta(\x)<\eta(\x'),f(\x)=f(\x')}[\eta(\x')-\eta(\x)]\\
&=&E_{(\eta(\x)-\eta(\x'))(f(\x)-f(\x'))<0}[|\eta(\x)-\eta(\x')|] +\tfrac{1}{2}E_{f(\x)=f(\x')}[|\eta(\x')-\eta(\x)|]\\
&\leq& E_{(\eta(\x)-\eta(\x'))(f(\x)-f(\x'))\leq0}[|\eta(\x)-\eta(\x')|],
\end{eqnarray*}
which yields that, from our assumption,
\begin{multline*}
R(f)-R^* \leq E_{(\eta(\x)-\eta(\x'))(f(\x)-f(\x'))\leq0} [\kappa_0\big(C(\eta(\x),\eta(\x'),0) \\ -C(\eta(\x),\eta(\x'),f^*(\x)-f^*(\x'))\big)^{\kappa_1}].
\end{multline*}
By using the Jensen's inequality, we further obtain
\begin{multline*}
R(f)-R^*\leq \kappa_0\big(E_{(\eta(\x)-\eta(\x'))(f(\x)-f(\x'))\leq0}[C(\eta(\x),\eta(\x'),0) \\ -C(\eta(\x),\eta(\x'),f^*(\x)-f^*(\x'))]\big)^{\kappa_1}
\end{multline*}
for $0<\kappa_1<1$. This remains to prove that
\begin{eqnarray*}
\lefteqn{E[C(\eta(\x),\eta(\x'),0) -C(\eta(\x),\eta(\x'),f^*(\x)-f^*(\x'))] \leq R_\phi(f)-R_\phi^*}\\
&&=E[C(\eta(\x),\eta(\x'),f(\x)-f(\x'))-C(\eta(\x),\eta(\x'),f^*(\x)-f^*(\x'))]
\end{eqnarray*}
where the expectations take over $(\eta(\x)-\eta(\x')) (f(\x)-f(\x'))\leq0$. To see it, we consider the following cases:
\begin{itemize}
\item If $\eta(\x)=\eta(\x')$ then $C(\eta(\x),\eta(\x'),0)\leq C(\eta(\x),\eta(\x'),f(\x)-f(\x'))$ since $\phi$ is convex;
\item  If $f(\x)=f(\x')$ then $C(\eta(\x),\eta(\x'),0)= C(\eta(\x),\eta(\x'),f(\x)-f(\x'))$;
\item If $(\eta(\x)-\eta(\x'))(f(\x)-f(\x'))<0$, then $(f(\x)-f(\x'))(f^*(\x)-f^*(\x'))<0$ from the assumption $(f^*(\x)-f^*(\x'))(\eta(\x)-\eta(\x'))>0$. Thus, $0$ is between $f(\x)-f(\x')$ and $f^*(\x)-f^*(\x')$, and for convex function $\phi$, we have
    \begin{multline*}
    C(\eta(\x),\eta(\x'),0)\leq\max(C(\eta(\x),\eta(\x'),f(\x)-f(\x')),\\
    C(\eta(\x),\eta(\x'),f^*(\x)-f^*(\x')))=C(\eta(\x),\eta(\x'),f(\x)-f(\x')).
    \end{multline*}
\end{itemize}
This theorem follows as desired.\qed

\subsection{Proof of Corollary~\ref{coro:exp}}\label{sec:pf:app1}

For exponential loss $\phi(t)=e^{-t}$, we have the optimal function $f^*$ such that
\begin{equation}\label{eq:tt1}
f^*(\x)-f^*(\x')=\frac{1}{2}\ln\frac{\eta(\x)(1-\eta(\x'))}{\eta(\x')(1-\eta(\x)}
\end{equation}
by minimizing the conditional risk $C(\eta(\x),\eta(\x'),f(\x)-f(\x'))$, and this follows
\[
(f^*(\x)-f^*(\x'))(\eta(\x)-\eta(\x'))>0 \text{ for } \eta(\x)\neq\eta(\x').
\]
From Eqn.~\eqref{eq:tt1}, we have
\[
C(\eta(\x),\eta(\x'),f^*(\x)-f^*(\x'))=\tfrac{1}{p(1-p)}\sqrt{\eta(\x)\eta(\x') (1-\eta(\x'))(1-\eta(\x))},
\]
and it is easy to get $C(\eta(\x),\eta(\x'),0)=\eta(\x)(1-\eta(\x'))+ \eta(\x')(1-\eta(\x))$. Therefore, we have
\begin{eqnarray*}
\lefteqn{C(\eta(\x),\eta(\x'),0) -C(\eta(\x),\eta(\x'),f^*(\x)-f^*(\x'))}\\
&&=\frac{1}{2p(1-p)}\big(\sqrt{\eta(\x)(1-\eta(\x'))} -\sqrt{\eta(\x')(1-\eta(\x))}\big)^2\\
&&=\frac{1}{2p(1-p)}\frac{|\eta(\x)-\eta(\x')|^2}{(\sqrt{\eta(\x)(1-\eta(\x'))} +\sqrt{\eta(\x')(1-\eta(\x))})^2}\\
&&\geq |\eta(\x)-\eta(\x')|^2/(2p(1-p)),
\end{eqnarray*}
where the last inequality holds from $\eta(x),\eta(x')\in [0,1]$. Hence, this lemma holds by applying Theorem~\ref{thm:con} to exponential loss.\qed

\subsection{Proof of Corollary~\ref{coro:logistic}}\label{sec:pf:app2}

For logistic loss $\phi(t)=\ln(1+e^{-t})$, we have the optimal function $f^*$ such that
\begin{equation}\label{eq:tt2}
f^*(\x)-f^*(\x')=\ln\frac{\eta(\x)(1-\eta(\x'))}{\eta(\x')(1-\eta(\x)},
\end{equation}
by minimizing the conditional risk $C(\eta(\x),\eta(\x'),f(\x)-f(\x'))$, and this immediately yields
\[
(f^*(\x)-f^*(\x'))(\eta(\x)-\eta(\x'))>0 \text{ for }\eta(\x)\neq\eta(\x').
\]
Therefore, we complete the proof by applying Theorem~\ref{thm:con} to logistic loss if the following holds:
\begin{multline}\label{eq:tt3}
C(\eta(\x),\eta(\x'),0) -C(\eta(\x),\eta(\x'),f^*(\x)-f^*(\x'))\\
\geq |\eta(\x)-\eta(\x')|^2/(8p(1-p)).
\end{multline}

We will prove that Eqn.~\eqref{eq:tt3} holds for $|\eta(\x')-0.5|\leq|\eta(\x)-0.5|$, and similar derivation could be made when $|\eta(\x')-0.5|>|\eta(\x)-0.5|$. For simplicity, we denote by $\eta=\eta(\x)$ and $\eta'=\eta(\x')$. Fix $\eta'$ and we set
\[
F(\eta)=2p(1-p)\big(C(\eta,\eta',0) -C(\eta,\eta',f^*(x)-f^*(x'))- (\eta-\eta')^2/(8p(1-p))\big).
\]
From Eqn.~\eqref{eq:tt2}, we further get
\begin{eqnarray*}
F(\eta)&=&\ln(2)(\eta+\eta' -2\eta'\eta)-(\eta-\eta')^2/4 \\
 &&-\eta(1-\eta')\ln\Big(1+\frac{\eta'(1-\eta)}{\eta(1-\eta')}\Big) -\eta'(1-\eta)\ln\Big(1+\frac{\eta(1-\eta')}{\eta'(1-\eta)}\Big).
\end{eqnarray*}
It is easy to obtain $F(\eta')=0$ and the derivative
\begin{eqnarray*}
F'(\eta)&=&\ln(2)(1-2\eta')-(\eta-\eta')/2 \\
&&-(1-\eta') \ln\Big(1+\frac{\eta'(1-\eta)}{\eta(1-\eta')}\Big) + \eta'\ln\Big(1+\frac{\eta(1-\eta')}{\eta'(1-\eta)}\Big).
\end{eqnarray*}
Further, we have $F'(\eta')=0$ and the second-order derivative
\[
F''(\eta)=\frac{\eta'(1-\eta')}{\eta(1-\eta)(\eta+\eta'-2\eta\eta')}-\frac{1}{2}\geq0,
\]
where the inequality holds since $\eta+\eta'-2\eta\eta'=\eta(1-\eta')+\eta'(1-\eta)<2$ and $\eta'(1-\eta')\geq \eta(1-\eta)$ from assumption $|\eta'-0.5|\leq|\eta-0.5|$. Therefore, $F'(\eta)$ is a non-decreasing function, and this yields that
\[
F'(\eta)\leq F'(\eta')=0 \text{ for }\eta\leq \eta',\text{ and }F'(\eta)\geq F'(\eta')=0\text{ for }\eta\geq \eta',
\]
which implies that $F(\eta)\geq F(\eta')=0$. Therefore, we complete the proof.\qed

\subsection{Proofs of Theorem~\ref{thm:acc:auc}}\label{pf:thm:acc:auc}

For accuracy's exponential surrogate loss, we have
\[
R_{\phi_\text{acc}}(f)-R^*_{\phi_\text{acc}}= E_{\x} \Big(\sqrt{\eta(\x)e^{-f(\x)}} -\sqrt{(1-\eta(\x))e^{f(\x)}}\Big)^2,
\]
and for AUC's exponential surrogate loss, we have
\begin{eqnarray*}
2p(1-p)(R_{\phi}(f)-R^*_{\phi})&=&E_{\x,\x'}\Big[ \Big(\sqrt{\eta(\x) (1-\eta(\x')) e^{-f(\x)+f(\x')}}\\
&&~~~~~~~~~~-\sqrt{\eta(\x')(1-\eta(\x))e^{f(\x)-f(\x')}}\Big)^2\Big].
\end{eqnarray*}
By using the fact $(ab-cd)^2\leq a^2(b-d)^2+d^2(a-c)^2$, it holds that
\begin{eqnarray*}
\lefteqn{2p(1-p)(R_{\phi}(f)-R^*_{\phi})}\\
&\leq&2E_{\x'}[(1-\eta(\x'))e^{f(\x)}]E_{\x}\Big[ \Big(\sqrt{\eta(\x)e^{-f(\x)}}-\sqrt{(1-\eta(\x))e^{f(\x)}}\Big)^2\Big]\\
&&+2E_{\x}[(1-\eta(\x))e^{f(\x)}]E_{\x'}\Big[\Big(\sqrt{(1-\eta(\x')) e^{f(\x')}}-\sqrt{\eta(\x')e^{-f(\x')}}\Big)^2\Big]\\
&=&4E_{\x}[(1-\eta(\x))e^{f(\x)}] (R_{\phi_\text{acc}}(f)-R^*_{\phi_\text{acc}}),
\end{eqnarray*}
and in a similar manner, we have
\[
2p(1-p)(R_{\phi}(f)-R^*_{\phi})\leq 4E_{\x}[\eta(\x)e^{-f(\x)}] (R_{\phi_\text{acc}}(f) -R^*_{\phi_\text{acc}}).
\]
This follows
\begin{eqnarray*}
\lefteqn{p(1-p)(R_{\phi}(f)-R^*_{\phi})}\\
&\leq&E_{\x}[\eta(\x)e^{-f(\x)}+(1-\eta(\x))e^{f(\x)}] (R_{\phi_\text{acc}}(f) -R^*_{\phi_\text{acc}})\\
&=&R_{\phi_\text{acc}}(f) (R_{\phi_\text{acc}}(f) -R^*_{\phi_\text{acc}})
\end{eqnarray*}
which completes the proof.\qed

\subsection{Proofs of Theorem~\ref{thm:auc:acc}}\label{pf:thm:auc:acc}

For a score function $f(\x)$, we have
\begin{eqnarray*}
\lefteqn{R_{\phi_\text{acc}}(f-t^*_f)-R_{\phi_\text{acc}}^*}\\
&=& E_{\x}[\eta(\x)e^{-f(\x)+t^*_f}+(1-\eta(\x))e^{f(\x)-t^*_f}] -2E_{\x}\sqrt{\eta(\x)(1-\eta(\x))}\\
&=&2\sqrt{E_{\x}[\eta(\x)e^{-f(\x)}]E_{\x}[(1-\eta(\x))e^{f(\x)}]} -2E_{\x}\sqrt{\eta(\x)(1-\eta(\x))}
\end{eqnarray*}
where the last equality holds from
\[
t_f^*=\tfrac{1}{2}\ln E_\x[\eta(\x)e^{-f(\x)}]- \tfrac{1}{2}\ln E_\x[(1-\eta(\x))e^{f(\x)}].
\]
For pairwise exponential loss of AUC, we have
\begin{eqnarray*}
\lefteqn{2p(1-p)(R_{\phi}(f)-R^*_{\phi})} \\
&=&E_{\x,\x'}[\eta(\x)(1-\eta(\x'))e^{f(\x')-f(\x)}+ \eta(\x')(1-\eta(\x))e^{f(\x)-f(\x')}]\\
&&-2E_{\x,\x'}[\sqrt{\eta(\x)\eta(\x')(1-\eta(\x))(1-\eta(\x'))}]\\
&=&2E_{\x}[\eta(\x)e^{-f(\x)}]E_{\x}[1-\eta(\x)e^{f(\x)}] -2(E_{\x}[\sqrt{\eta(\x)(1-\eta(\x))}])^2\\
&\geq&2\left(\sqrt{E_{\x}[\eta(\x)e^{-f(\x)}]E_{\x}[(1-\eta(\x))e^{f(\x)}]} -2E_{\x}\sqrt{\eta(\x)(1-\eta(\x))}\right)^2\\
&=&\tfrac{1}{2}(R_{\phi_\text{acc}}(f-t^*_f)-R_{\phi_\text{acc}}^*)^2
\end{eqnarray*}
which completes the proof.\qed

\section{Conclusion and Open Problems}\label{sec:com}

AUC (area under ROC curve) is a popular evaluation criterion widely used in diverse learning tasks. Many learning approaches are developed, and most work with pairwise surrogate losses owing to the non-convexity and discontinuousness of AUC. Therefore, it is important to study the consistency of learning algorithms based on minimizing pairwise surrogate losses.

We first showed that calibration is necessary yet insufficient for AUC consistency, e.g., hinge loss and absolute loss are calibrated but inconsistent with AUC. Based on this finding, we provide a new sufficient condition for the asymptotic consistency of learning approaches based on surrogate loss functions, and many surrogate losses such as exponential loss, logistic loss, least-square hinge loss, etc., are proven to be consistent. We also derive the regret bounds for exponential loss and logistic loss, and obtain the regret bounds for many surrogate losses under the realizable setting. Finally, we provide regret bounds to show the equivalence between the exponential surrogate loss of AUC and exponential surrogate loss of accuracy, and one straightforward consequence of such finding is that AdaBoost and RankBoost are equivalent in the limit of infinite sample.

It is worth mentioning that our theoretical study has already inspired the design of new algorithms. For example, by optimizing the pairwise least square loss, \cite{Gao:Jin:Zhu:Zhou2013} proposed the OPAUC algorithm which requires only one scan of data to optimize AUC, while its performance is superior to previous AUC optimization algorithms that optimizes hinge loss.

\bibliographystyle{elsarticle-harv}
\bibliography{reference}
\end{document}